\documentclass{article}

     \PassOptionsToPackage{numbers, compress}{natbib}

 \usepackage[preprint]{tackling_climate_workshop_style}

\usepackage[utf8]{inputenc} 
\usepackage[T1]{fontenc}    
\usepackage{hyperref}       
\usepackage{url}            
\usepackage{booktabs}       
\usepackage{amsfonts}       
\usepackage{nicefrac}       
\usepackage{microtype}      
\usepackage{graphicx}
\usepackage{amsmath}
\usepackage{float}
\usepackage{color}
\usepackage{lineno}
\usepackage{comment}
\usepackage{subcaption}

\title{SoftSEEPS improves ML-based precipitation forecasting}

\author{%
  Jost Arndt\thanks{These authors contributed equally to this work.} \\
  Fraunhofer Heinrich Hertz Institute \\
  Berlin, Germany \\
  \texttt{jost.arndt@hhi.fraunhofer.de} \\
  \And
  Utku Isil\footnotemark[1]\\
  Fraunhofer Heinrich Hertz Institute \\
  Berlin, Germany \\
  \And
  Noelia Otero \\
  Fraunhofer Heinrich Hertz Institute \\
  Berlin, Germany \\
  \And
  Rodrigo Almeida \\
  Fraunhofer Heinrich Hertz Institute \\
  Berlin, Germany \\
  \AND
  Wojciech Samek \\
  Fraunhofer Heinrich Hertz Institute \\
  Technische Universität Berlin \\
  BIFOLD \\ 
  Berlin, Germany \\
  \And
  Jackie Ma \\
  Fraunhofer Heinrich Hertz Institute \\
  Berlin, Germany \\
}

\begin{document}

\maketitle

\begin{abstract}
In this paper we have developed a differentiable approximation of the well-known SEEPS score, which we name \textit{SoftSEEPS}. This allows the training of a Machine Learning model to forecast precipitation directly. We test \textit{SoftSEEPS} on the IMERG dataset ($0.1^\circ$ resolution) by training a decoder for precipitation on the latent space of a pre-trained low-resolution forecasting model. Combining \textit{SoftSEEPS} and RMSE in a joint objective is possible with marginal trade-offs in either metric.
\end{abstract}

\section{Introduction}
\label{sec:intro}

Heavy precipitation is intensifying with climate change~\cite{ipcc_ar6_wg1}, and early-warning and disaster-response systems that rely on precipitation forecasts care primarily about getting the dry/light/heavy category right, not intensity, making the training/evaluation metric itself a climate-impact question. Machine-learning (ML)-based weather forecasting has recently surpassed traditional numerical models on several benchmark scores~\cite{pangu, weatherMesh3, graphcast, fourcastnet, weatherbench, weatherbench_2}. However, forecasting precipitation, remains challenging: unlike pressure or temperature~\cite{weatherbench, weatherbench_2}, its distribution is regionally variable, intermittent, and strongly skewed~\cite{rainbench, Sun2014-de, Ebert-Uphoff2023-eq}.
Standard error metrics such as root mean squared error (RMSE) poorly reflect forecast quality for precipitation at high spatial resolutions, which are needed for better climate adaptation and disaster response. Therefore, the stable equitable error in probability space (SEEPS)~\cite{seeps_original, seeps_assessment, weatherbench_2} instead is often used for evaluation. It scores forecasts categorically into \textit{dry}, \textit{light precipitation}, or \textit{heavy precipitation} using climatological thresholds estimated separately at each location. Miscategorization gets penalized by a climatology-dependent weight. This categorization is a piecewise-constant assignment, so SEEPS provides no gradient signal, thus ML forecasting models can only be evaluated against it, but not directly optimized for it.

Even though it is argued that RMSE should not be used for training~
\cite{stopRMSE}, none of the papers in~\citet{weatherbench} have developed a differentiable approximation of the SEEPS score~\citep{pangu, graphcast, fuxi, fuxi2, GenCast, weatherMesh3}. We close this gap with a differentiable, sigmoid-relaxed approximation of SEEPS that can be used directly as a training loss and call it \textit{SoftSEEPS}. 
Prior differentiable relaxations of categorical operations, such as the Gumbel-softmax and concrete distributions~\cite{jang2017gumbel, maddison2017concrete} have shown good success across different ML domains and motivate our sigmoid-based relaxation.

\section{Methods}
\subsection{SoftSEEPS}
\label{sec:diff_seeps}
\begin{figure}[H] 
\centering
\includegraphics[width=0.95\textwidth]{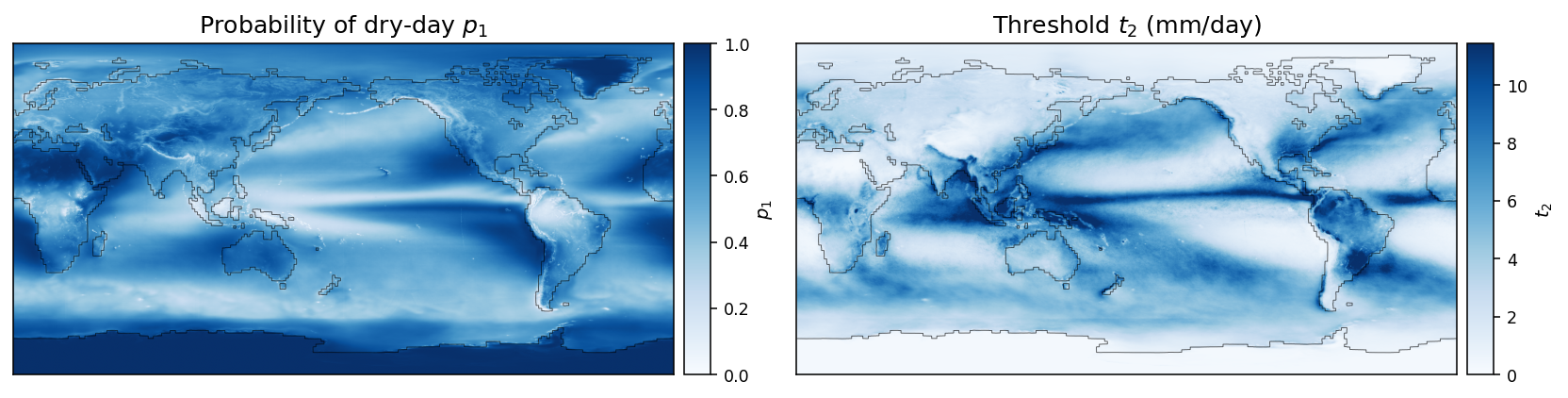}
    \caption{Geographic distribution of the annual-mean "dry precipitation"-probability $p_1$ and threshold values between light and heavy precipitation $t_2$.}
\label{fig:clim_annual_summary}
\end{figure}

The original SEEPS score classifies both the forecast and the ground truth into the categories \emph{dry}, \emph{light precipitation}, and \emph{heavy precipitation}, using localized historical climatological thresholds $t_1, t_2$. While $t_1=0.25mm$ is constant, $t_2$ is set as the threshold dividing two-thirds of the mass(given that it rains) for day of year, since typical precipitation varies strongly with region and season (e.g. rainforest vs. desert, summer vs. winter). The joint category pairs of forecast/observed are recorded in a $3\times3$ contingency table and scored with a location and day-of-year-dependent penalty matrix $S$ that weights the table entrywise before summing, i.e.\ a Frobenius inner product. The penalty matrix $S$ has, for a specified location and day-of-year, the form
\begin{linenomath*}
$$
S =  \frac{1}{2}
\begin{pmatrix}
0 & \frac{1}{1-p_1} & \frac{1}{p_3}+\frac{1}{1-p_1} \\[2pt]
\frac{1}{p_1} & 0 & \frac{1}{p_3} \\[2pt]
\frac{1}{p_1}+\frac{1}{1-p_3} & \frac{1}{1-p_3} & 0
\end{pmatrix},
$$
\end{linenomath*}
where rows index the forecast category and columns the observed category, both ordered dry, light, heavy. $p_1$ is the dry-day probability for a given location and day-of-year, and $p_3$ the heavy-precipitation probability. By construction $p_3 = (1-p_1)/3$, since the wet-day mass is split 2:1 between the light and heavy categories. The annual means of both the $p_1$ probability and $t_1$ thresholds can be found in Fig~\ref{fig:clim_annual_summary}.
The exact content of $S$ is derived and justified in~\cite{seeps_original}. A perfect forecast scores $0$; SEEPS is \emph{equitable}, so an unskilled (e.g.\ climatological) forecast has an expected score of $1$, giving a natural skill reference point.

\begin{figure}[H]
    \centering
    \includegraphics[width=0.42\linewidth]{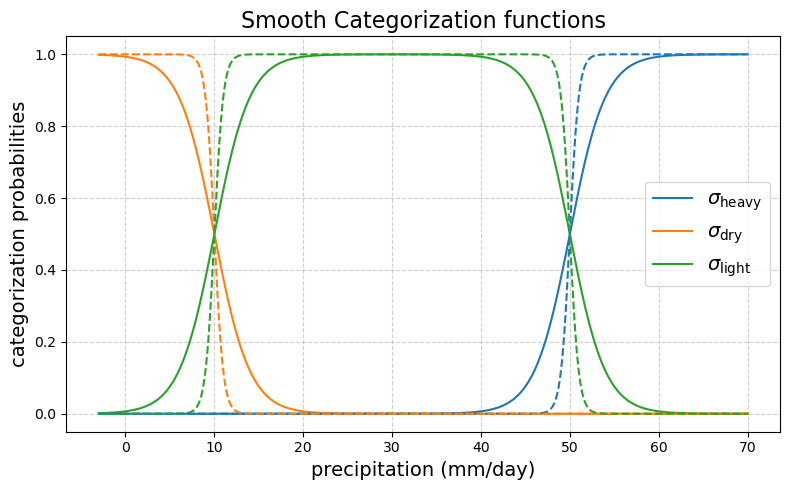}
    \caption{Smooth categorization functions $c(x,\tau)$: dashed lines use a smoothing parameter $\tau=0.5$, solid lines $\tau=2$.}
    \label{fig:soft_categories}
\end{figure}

To enable gradient-based optimization based on SEEPS, we  approximate the hard categorical indicators with a differentiable "soft classification" function based on sigmoid-relaxed category probabilities, we define the classificatin function
\begin{linenomath*}
$$
c(x, \tau) :
=
\begin{bmatrix}
p_{\mathrm{dry}}(x, \tau)\\
p_{\mathrm{light}}(x, \tau)\\
p_{\mathrm{heavy}}(x, \tau)
\end{bmatrix},
$$
where $p_{\mathrm{dry}}(x, \tau) = \sigma\!\left(\frac{t_1 - x}{\tau}\right)$, $p_{\mathrm{heavy}}(x, \tau) = \sigma\!\left(\frac{x - t_2}{\tau}\right)$, with $\sigma(\cdot)$ being the sigmoid function, $\tau > 0$ a smoothing parameter, $x$ the forecast precipitation in physical units (mm/day). $p_{\mathrm{light}}(x, \tau) := 1 - p_{\mathrm{dry}}(x, \tau) - p_{\mathrm{heavy}}(x, \tau)$ so the vector sums to $1$. These soft categorization functions are shown in Fig.~\ref{fig:soft_categories}. $c(x,\tau)$ is differentiable in $x$ and converges to the original hard classification for $\tau \to 0$.  Higher smoothings trade off approximation accuracy for better-conditioned gradients, which is what makes $c(x,\tau)$ usable as a training signal. Combining $c(x,\tau)$ with the location- and day-of-year-wise penalty matrix $S$, lets us define the \textit{SoftSEEPS} score between a prediction $x$ and ground truth $y$ as
$$
s(x, y, \tau )  := c(x, \tau)^T S \, c(y, \tau).
$$
\end{linenomath*}
This is a fully differentiable sigmoid relaxation of the categorical assignment, governed by a smoothing parameter, in the same spirit as other differentiable relaxations of categorical operations used in ML~\cite{jang2017gumbel, maddison2017concrete}. We succesfully have tested that \textit{SoftSEEPS} converges to reference discrete-SEEPS implementation from the \texttt{scores} library~\cite{scores_library}. Rather than fixing $\tau$, we anneal it automatically during training with a plateau scheduler that tracks validation discrete SEEPS and lowers $\tau$ once it plateaus. While we use \textit{SoftSEEPS} for training, all reported validation-SEEPS are the discrete scores.

\subsection{Model: High-resolution Precipitation Decoder}
To study \textit{SoftSEEPS} in a high-resolution forecasting setting, we use $1.5^{\circ}$ resolution ERA5~\cite{era5} as input and GPM IMERG~\cite{imerg} Final daily precipitation (V07B, mm/day, $0.1^{\circ}$ resolution) as the training target, shifted by 24 hours as a forecasting-task. We train only a convolutional precipitation decoder (IMERGDecoder) that maps the latent space of a publicly available, frozen ArchesWeather-M~\cite{arches} backbone to the precipitation target. The model predicts next-day (24-hour accumulated) IMERG precipitation: the backbone latent is decoded to IMERG resolution without any explicit precipitation input. The 54.3\,M decoder parameters are the sole trainable parameters. This setup lets us combine a well-performing yet inexpensive weather forecasting model with a limited training-compute budget.

\textbf{IMERGDecoder} is a convolutional upsampling network that maps a backbone latent of size $384 \times 60 \times 120$ to IMERG resolution ($1 \times 1800 \times 3600$) through three bilinear-upsampling stages ($\times2$, $\times3$, $\times5$). Thus, although the backbone operates on a substantially coarser spatial grid, the decoder produces precipitation directly on the $0.1^{\circ}$ resolution IMERG grid. Each stage applies four residual blocks before upsampling, using 512 channels in the first two stages and 384 in the third.

\textbf{Baselines.} Direct comparison to existing learned downscaling methods is challenging, as prior work at $0.1^{\circ}$ typically focuses on regional or patch-based  settings~\cite{RainShift}. We therefore use persistence (the previous day's IMERG) and climatology (the locationwise, day-of-year-wise mean IMERG over the training set) as reference baselines. Additionally, we compare against bicubic interpolation of same-day ERA5 total precipitation to 
$0.1^{\circ}$.

\section{Experiments}
\label{sec:setup}

For training we use the years 1998--2018, validation 2019, and testing 2020. To test the impact of \textit{SoftSEEPS} as optimization objective we compare three settings: MSE with log normalized data only ($\mathrm{mean}[(\log(1{+}\hat{y}) - \log(1{+}y))^2]$),  \textit{SoftSEEPS}-only; and the linear combination of both, which we simply denote as  MSE $+ \lambda \times$ \textit{SoftSEEPS} with different $\lambda \in \{ 0.1, 0.25, 0.5, 1, 1.5, 1.75 \}$.
We then train the model for 25,000 steps using the AdamW optimizer on a single A100 GPU, lasting approximately 5 hours wall-clock time.

\section{Results}
\label{sec:results}
\begin{figure}[H]
\centering
\includegraphics[width=0.95\textwidth]{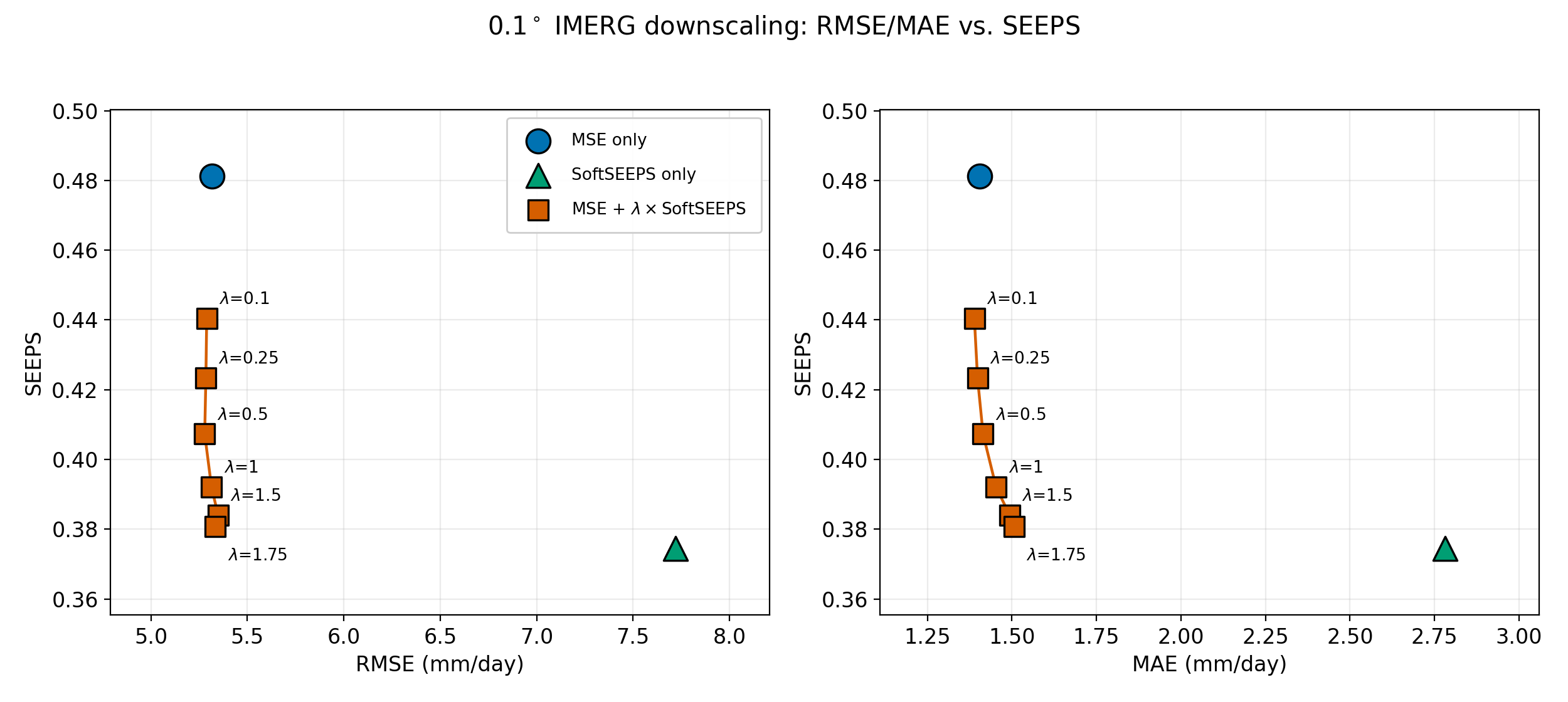}

\caption{$0.1^\circ$ IMERG test-set RMSE (left) and MAE (right) vs.\ SEEPS. Circle/triangle: MSE-only and \textit{SoftSEEPS}-only (Table~\ref{tab:results}). Squares: the combined loss MSE $+\lambda\times$\textit{SoftSEEPS} swept over $\lambda\in\{0.1,0.25,0.5,1.0,1.5,1.75\}$.}
\label{fig:tradeoff_paper}
\end{figure}

Training the proposed IMERGDecoder \textit{SoftSEEPS} gives converging training error. Validating on a proper SEEPS implementation shows that the learned objective is meaningful for SEEPS as well. Further, by sweeping through multiple values for $\lambda$, we can see that optimization on a linear combination of \textit{SoftSEEPS} and log normalized RMSE is possible with only marginal trade-offs, as shown in Fig.~\ref{fig:tradeoff_paper}.

A natural expectation is that \textit{SoftSEEPS} can only trade off against RMSE. Fig.~\ref{fig:tradeoff_paper} shows that the $\lambda$ sweep instead appears to trace out a Pareto-optimal frontier between RMSE and SEEPS: SEEPS drops by as much as $18.51\%$ at no resolvable cost in RMSE. 

\begin{figure}[H]
\centering
\includegraphics[width=\textwidth]{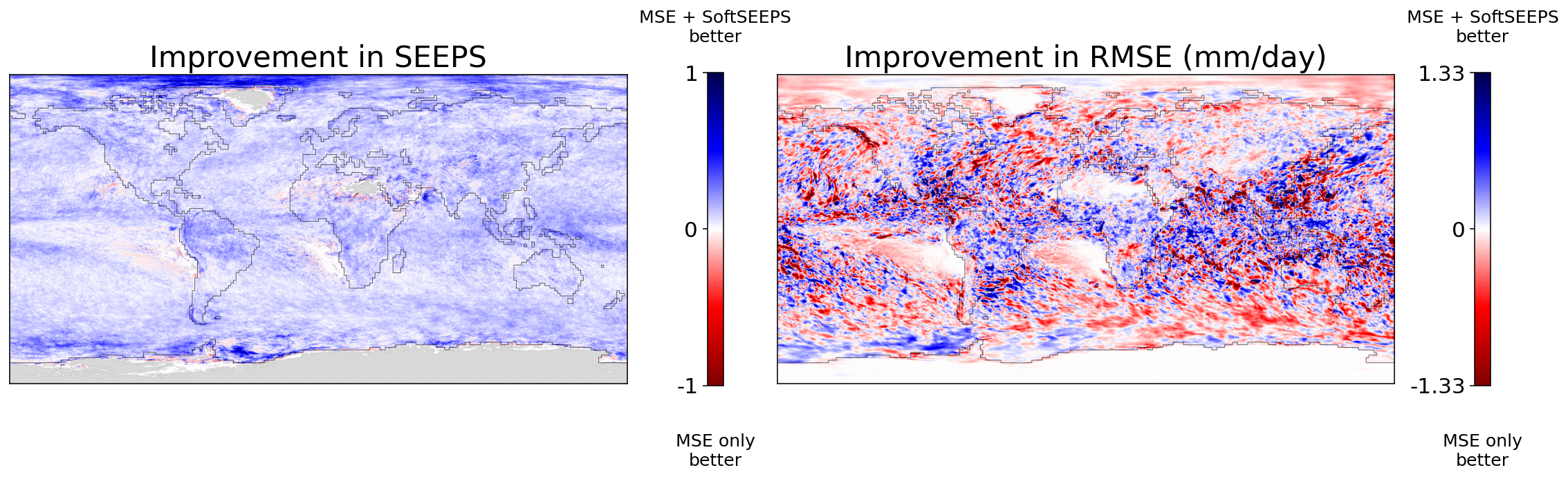}
\caption{Locationwise improvement from adding \textit{SoftSEEPS} to the training loss, for SEEPS (left) and RMSE (right), computed as difference between the evaluation metrics. Blue indicates adding \textit{SoftSEEPS} is better at that location and in that metric; red indicates MSE only is better.}
\label{fig:geo_comparison}
\end{figure}

Beyond this aggregate trade-off, we also study where on the globe adding \textit{SoftSEEPS} to the loss helps or hurts. Fig.~\ref{fig:geo_comparison} shows that evaluation in SEEPS improves spatially uniformly, while evaluation in RMSE both improves and degrades performance in a spatially varying pattern. We see this as an indicator that the inclusion of climatological information in the training objective is successful and helps the models. Furthermore, the decoder trained on the original metric already exceeds the baselines, which can be seen in the Appendix in Table~\ref{tab:results}.

\section{Conclusion}
\label{sec:conclusion}
We have demonstrated, that the utilization of our \textit{SoftSEEPS} improves precipitation forecasting on discrete SEEPS score, which is a commonly respected metric in weather forecasting. We have further demonstrated that training a linear combination of \textit{SoftSEEPS} and MSE is possible with overall a marginal trade-off. Practically, this lets precipitation models be trained directly for the dry/light/heavy distinctions that early-warning and disaster-response systems already act on. Future works should train larger weather models for precipitation on \textit{SoftSEEPS}. Ultimately, scaling these methods provides a clear pathway to climate impact, enhancing weather forecasting to better mitigate and adapt to the severe precipitation events driven by climate change.

\newpage

\begin{ack}
\end{ack}

\bibliographystyle{abbrvnat}
\bibliography{Literature}

\section{Appendix}
\label{sec:appendix}

\begin{table}[H]
\centering
\caption{Next-day precipitation forecasting at $0.1^\circ$ (IMERG, test year 2020).}
\label{tab:results}
\begin{tabular}{lccc}
\toprule
\textbf{Setting} & \textbf{RMSE} $\downarrow$ (mm/day) & \textbf{MAE} $\downarrow$ (mm/day) & \textbf{SEEPS} $\downarrow$ \\
\midrule
\multicolumn{4}{l}{\textit{Baselines}} \\
\quad Persistence                      & 9.288 & 2.927 & 0.819 \\
\quad Climatology                      & 7.492 & 2.853 & 1.098 \\
\quad ERA5 interpolation            & 5.689 & 1.951 & 0.612 \\
\midrule
\multicolumn{4}{l}{\textit{Arches+Decoder}} \\
\quad MSE                                    & \textbf{5.316} & \textbf{1.404} & 0.481 \\
\quad \textit{SoftSEEPS} only                             & 7.720 & 2.781 & \textbf{0.375} \\
\quad MSE $+1.75 \times$\textit{SoftSEEPS}                 & \underline{5.334} & \underline{1.507} & \underline{0.381} \\
\bottomrule
\end{tabular}
\end{table}

\begin{figure}[H]
\centering
\includegraphics[width=0.85\textwidth]{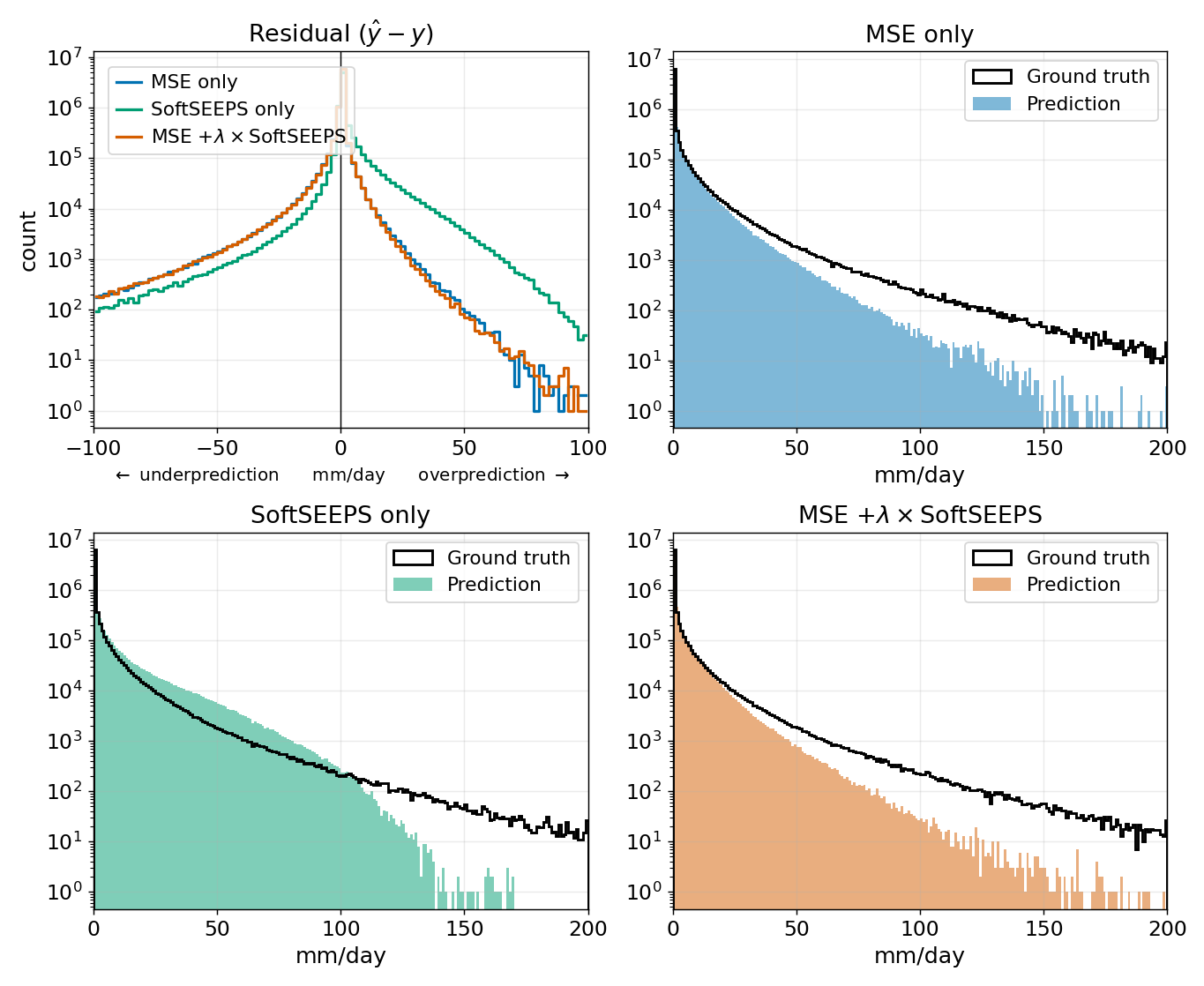}
\caption{Test-set residual ($\hat{y}-y$, top left, three objectives overlaid) and predicted vs.\ ground-truth precipitation distribution (remaining panels, one per objective), all log-scale.}
\label{fig:residual_hist}
\end{figure}

\end{document}